\documentclass[letterpaper]{article} 
\usepackage{aaai23}  
\usepackage{times}  
\usepackage{helvet}  
\usepackage{courier}  
\usepackage[hyphens]{url}  
\usepackage{graphicx} 
\usepackage{natbib}  
\usepackage{caption} 
\usepackage{algorithm}
\usepackage{algorithmic}

\usepackage{amsmath}
\usepackage{multirow}

\usepackage{newfloat}
\usepackage{listings}
\DeclareCaptionStyle{ruled}{labelfont=normalfont,labelsep=colon,strut=off} 
\floatstyle{ruled}
\newfloat{listing}{tb}{lst}{}
\floatname{listing}{Listing}
\title{MRCN: A Novel Modality Restitution and Compensation Network for Visible-Infrared Person Re-identification}
\author{Yukang Zhang\textsuperscript{\rm 1}, Yan Yan\textsuperscript{\rm 1}, Jie Li\textsuperscript{\rm 2}, Hanzi Wang\textsuperscript{\rm 1, 3}\thanks{Corresponding author.}}

\affiliations{

     \textsuperscript{\rm 1}Fujian Key Laboratory of Sensing and Computing for Smart City, School of Informatics, Xiamen University, China.
          
     \textsuperscript{\rm 2}Video and Image Processing System Laboratory, School of Electronic Engineering, Xidian University, Xi’an, China.

     \textsuperscript{\rm 3}Shanghai Artificial Intelligence Laboratory, Shanghai, China.

     zhangyk@stu.xmu.edu.cn, \{yanyan, hanzi.wang\}@xmu.edu.cn, leejie@mail.xidian.edu.cn

}

\usepackage{bibentry}

\begin{document}

\maketitle

\begin{abstract}
Visible-infrared person re-identification (VI-ReID), which aims to search identities across different spectra, is a challenging task due to large cross-modality discrepancy between visible and infrared images. The key to reduce the discrepancy is to filter out identity-irrelevant interference and effectively learn modality-invariant person representations. In this paper, we propose a novel Modality Restitution and Compensation Network (MRCN) to narrow the gap between the two modalities. Specifically, we first reduce the modality discrepancy by using two Instance Normalization (IN) layers. Next, to reduce the influence of IN layers on removing discriminative information and to reduce modality differences, we propose a Modality Restitution Module (MRM) and a Modality Compensation Module (MCM) to respectively distill modality-irrelevant and modality-relevant features from the removed information. Then, the modality-irrelevant features are used to restitute to the normalized visible and infrared features, while the modality-relevant features are used to compensate for the features of the other modality. Furthermore, to better disentangle the modality-relevant features and the modality-irrelevant features, we propose a novel Center-Quadruplet Causal (CQC) loss to encourage the network to effectively learn the modality-relevant features and the modality-irrelevant features. Extensive experiments are conducted to validate the superiority of our method on the challenging SYSU-MM01 and RegDB datasets. More remarkably, our method achieves 95.1\% in terms of Rank-1 and 89.2\% in terms of mAP on the RegDB dataset.
\end{abstract}

\section{Introduction}

Person re-identification (person ReID) is attracting more and more attention due to its great application potential in intelligent surveillance systems. Given a query image, the goal of person ReID is to match the most relevant person in a non-overlapping camera surveillance system \cite{sun2018beyond, wang2018learning, zheng2019pyramidal, tan2022dynamic}. Existing person ReID methods mainly focus on solving challenges related to variations in human pose, background and illumination. The desirable performance of those methods largely depends on good visible light conditions to clearly catch the appearance characteristics of humans \cite{Kalayeh_2018_CVPR, tay2019aanet, hou2019interaction, Gao_2020_CVPR, Miao_2019_ICCV}. However, when the light conditions are not desirable, the surveillance system usually automatically switches from the visible (VIS) modality to the near-infrared (NIR) modality to cope with the problem of low illumination \cite{fu2019dual, fu2021dvg, fu2021high}. Therefore, it is necessary to consider the crucial problem of visible-infrared person ReID (VI-ReID). 
The goal of VI-ReID is to match persons captured by the VIS and NIR cameras with different spectra. Compared with the widely studied single-modality person ReID, the VI-ReID is much more challenging due to additional cross-modality discrepancy between the VIS and NIR images \cite{ye2020deep, zhang2021towards}.

\begin{figure}[t]
\includegraphics[height=4.4cm,width=8.2cm]{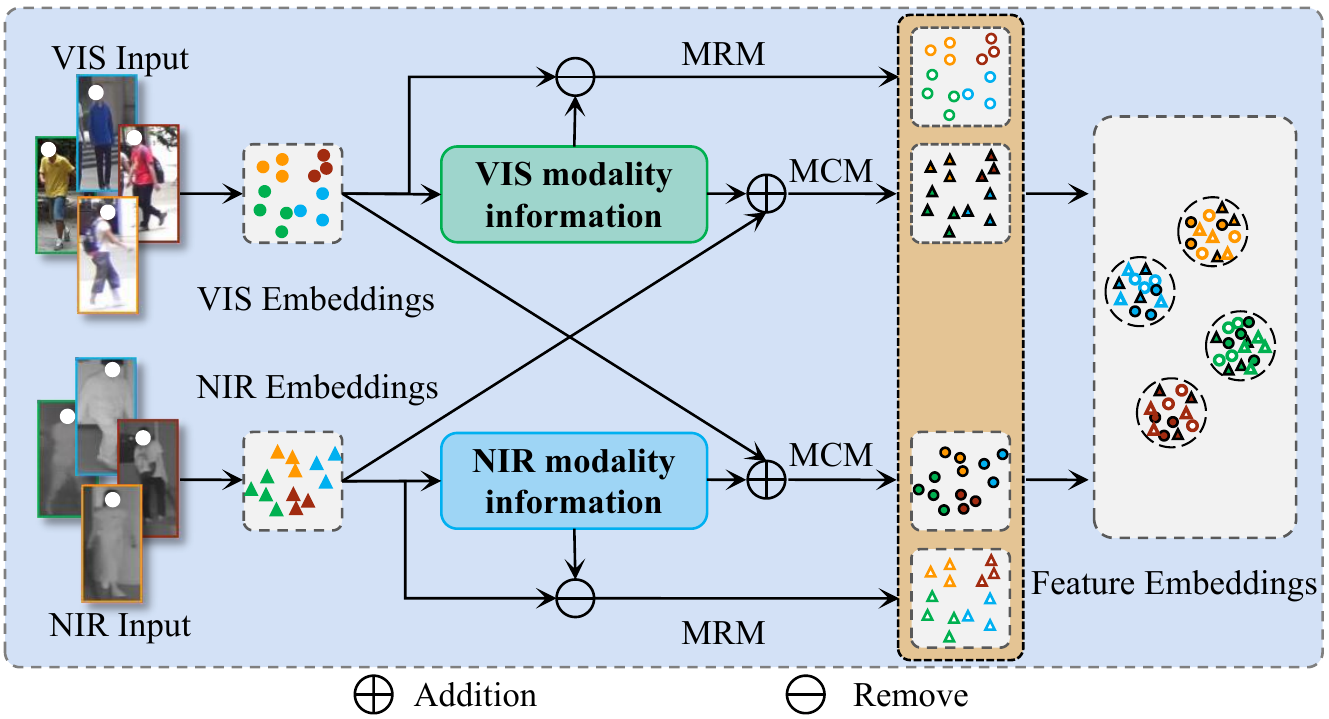}
\caption{Illustration of the proposed method. The person images captured from different modalities have large modality gap. To narrow this gap, we propose a novel MRCN to disentangle the modality-relevant and modality-irrelevant features and reduce the modality discrepancy between the VIS and NIR images.}
\label{img:img1}
\end{figure}

To reduce the modality discrepancy between the VIS and NIR images, two popular types of methods have been proposed. One type is the image-level methods~\cite{wang2019learning, wang2020cross, wang2019aligngan, choi2020hi}, which aim to translate a NIR (or VIS) image into its VIS (or NIR) counterpart by using the Generative Adversarial Networks (GANs)~\cite{goodfellow2014generative}. 
Despite their success in reducing modality discrepancy, generating cross-modality images is usually accompanied by noises due to the lack of VIS-NIR image pairs.
The other type is the feature-level methods \cite{wu2017rgb, dai2018cross, ye2020dynamic, wu2020rgb}, which typically train an end-to-end network to pull close the samples of the same identities with different modalities. Benefiting from the powerful feature extraction capability of Convolutional Neural Networks (CNNs), those methods have achieved good performance. However, the large modality discrepancy between the VIS and NIR images makes those methods difficult to project the cross-modality images into a common feature space directly. 

In this paper, we aim to minimize the modality discrepancy between the VIS and NIR images, which can be viewed as two types of images with different styles. The key is to find a way to separate the modality-relevant and modality-irrelevant information. Inspired by SNR \cite{jin2020style} in generalizing images of different styles, we propose a novel Modality Restitution and Compensation Network (MRCN) to reduce the modality discrepancy. The proposed MRCN eliminates the modality discrepancy between the VIS and NIR images by performing modality normalization on the VIS and NIR features through two Instance Normalization (IN) layers \cite{ulyanov2016instance}. After the two IN layers, two plug-and-play modules are proposed to further distill the modality-irrelevant and modality-relevant information. Then, the distilled modality-irrelevant information is restituted to the normalized features to ensure high performance of our model, while the distilled modality-relevant information is used to compensate for its normalized counterpart features to reduce the modality discrepancy between the VIS and NIR images. 
Thus, the proposed MRCN can effectively reduce the modality differences while ensuring high performance.

Furthermore, to better disentangle modality information and further reduce the modality discrepancy between the VIS and NIR images, we propose a novel Center-Quadruplet Causal loss (CQC) to encourage the network to effectively extract modality-relevant and modality-irrelevant information, which are respectively used to restitute modality-irrelevant information to the normalized modality features and compensate modality-relevant information for its counterpart modality features. 
With the incorporation of MRCN and the CQC loss into an end-to-end learning framework, the proposed method achieves an impressive performance on two challenging VI-ReID datasets.

Our contributions are summarized as follows:

$\bullet$ We propose a novel Modality Restitution and Compensation Network to disentangle the modality-relevant and modality-irrelevant features and reduce the modality discrepancy between the VIS and NIR images. In particular, the disentangled features can effectively reduce the modality discrepancy.

$\bullet$ We propose a Center-Quadruplet Causal loss to make the disentangled features consistent in modality distribution, which greatly facilitates the disentanglement of modality-relevant and modality-irrelevant features and significantly improves the performance of VI-ReID task.

$\bullet$ Extensive experiments with ablation studies show that the proposed MRCN outperforms several state-of-the-art methods on two challenging VI-ReID benchmarks. Specially, our method achieves 95.1\% in terms of Rank-1 and 89.2\% in terms of mAP on the RegDB dataset.

\begin{figure*}[t]
\centering
\includegraphics[height=7.2cm,width=17.8cm]{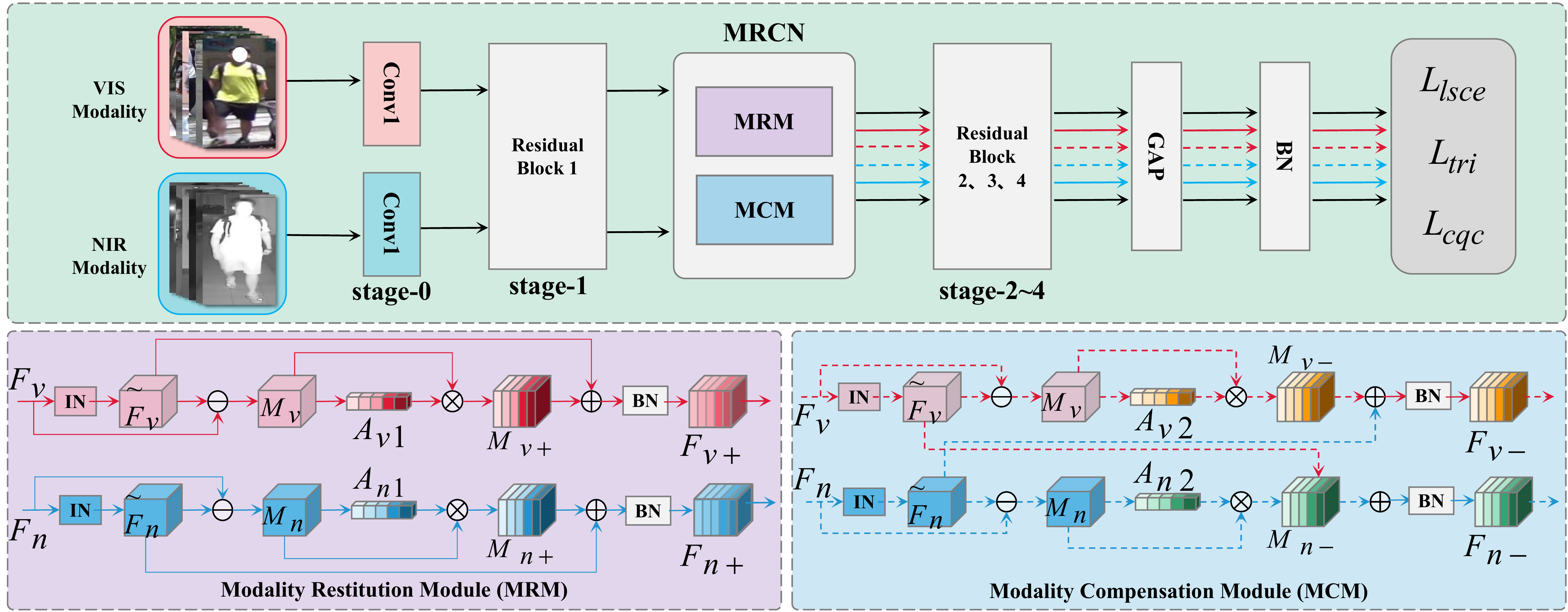}
\caption{The overall architecture of the proposed MRCN, including a Modality Restitution Module (MRM) and a Modality Compensation Module (MCM), which are respectively used to distill modality-irrelevant and modality-relevant information.}
\label{img:img2}
\end{figure*}

\section{Related Works}

There are two main categories of methods for the VI-ReID task: the image-level methods and the  feature-level methods.

\subsection{The Image-Level Methods}

The image-level methods reduce the modality discrepancy by transforming one modality into the other \cite{zhang2021towards}. For example, D$^{2}$RL \cite{wang2019learning} and AlignGAN \cite{wang2019aligngan} transform the NIR images into their VIS counterparts and transform the VIS images into their NIR counterparts. JSIA-ReID \cite{wang2020cross} generates cross-modality paired-images and it performs both global set-level and fine-grained instance-level alignments. Those methods often design complex generative models to align cross-modality images. Despite their success, generating cross-modality images is unavoidably accompanied by noises due to the lack of VIS-NIR image pairs. Recently, X-modality \cite{li2020infrared} and its variations (such as SMPL \cite{wei2021syncretic} and MMN \cite{zhang2021towards}) leverage a lightweight network to get an auxiliary modality to assist the cross-modality search. However, there is still a modality gap between this auxiliary modality and the VIS-NIR modality.

\subsection{The Feature-Level Methods}
The feature-level methods aim to find a modality-shared and modality-specific feature space, in which the modality discrepancy is minimized \cite{yang2022learning, zhang2022fmcnet, wu2021discover}. For this purpose, CM-NAS \cite{fu2021cm} utilizes a BN-oriented search space, where the standard optimization can be implemented. MCLNet \cite{hao2021cross} propose to minimize the inter-modality discrepancy while maximizing the cross-modality similarity. Inspired by adversarial learning, cmGAN \cite{dai2018cross} designs a cutting-edge discriminator to learn discriminative feature representations from different modalities. MPANet \cite{wu2021discover} introduce a modality alleviation module and a pattern alignment module to jointly extract discriminative features. However, the large modality discrepancy between the VIS and NIR images makes it difficult to project the cross-modality images into a common space directly. 

\section{Method}

\subsection{Model Architecture}

Fig. \ref{img:img2} provides an overview of the proposed MRCN. In MRCN, two Instance Normalization (IN) layers \cite{ulyanov2016instance} are first used to eliminate the modality discrepancy between the VIS and NIR modalities. Then, the proposed Modality Restitution Module (MRM) and Modality Compensation Module (MCM)) are used to reduce the influence of IN layers on removing discriminative information and reduce the modality gaps between the VIS and NIR images. Besides, we propose a new Center-Quadruplet Causal loss (CQC) to encourage the network to learn modality-relevant and modality-irrelevant features, which are respectively used to restitute modality-irrelevant information to the normalized modality features and compensate modality-relevant information for its counterpart modality features. During the inference process, the features extracted by MCM are not used. The original features and the features extracted by MRM are concatenated for testing.
\subsection{Modality Restitution and Compensation Network}
Due to the natural difference between the reflectivity of the VIS spectrum and the emissivity of the NIR spectrum, there is a large modality discrepancy between person images of different modalities. In this paper, we use two IN layers to perform modality normalization on the VIS and NIR modalities, respectively. However, the IN layers will lead to the loss of some discriminative information. Inspired by the work in SNR \cite{jin2020style}, we propose to further distill modality-irrelevant features from the removed information and restitute them to the network to ensure high discrimination of the extracted features. Moreover, we also distill modality-relevant features from the removed information and add them to its counterpart modality to compensate for the discrepancy between the two modalities. 

For convenience, we first define the VI-ReID task. The proposed MRCN takes an image pair of the same identities but different modalities as the input. Let $\textbf{F}_v$ and $\textbf{F}_n$ represent the features output by the first stage layer of the backbone corresponding to the VIS and NIR modalities, respectively. 
First, we use two IN layers to perform modality normalization for $\textbf{F}_v$ and $\textbf{F}_n$. 
For the VIS modality, we have:
\begin{small}
\begin{equation}
\widetilde{\textbf{F}}_v = IN(\textbf{F}_v) = \gamma_v (\frac{\textbf{F}_v -\mu (\textbf{F}_v)}{\sigma (\textbf{F}_v)} ) + \beta_v,
\end{equation}
\end{small}

and for the NIR modality, we have:
\begin{small}
\begin{equation}
\widetilde{\textbf{F}}_n = IN(\textbf{F}_n) = \gamma_n (\frac{\textbf{F}_n -\mu (\textbf{F}_n)}{\sigma (\textbf{F}_n)} ) + \beta_n, 
\end{equation}
\end{small}
where $\mu(\cdot)$ and $\sigma(\cdot)$ are respectively the mean and standard deviation of the features $\textbf{F}_v$ and $\textbf{F}_n$, which are computed across the spatial dimensions for each channel and each input \cite{jin2020style, huang2017arbitrary}. $\gamma_v$, $\gamma_n$, $\beta_v$ and $\beta_n$ are the parameters learned from the network. 

As the IN layers can filter out some modality information from the image content, the difference $\textbf{M}_v$ (or $\textbf{M}_n$) between the original features $\textbf{F}_v$ (or $\textbf{F}_n$) and the modality normalized features $\widetilde{\textbf{F}_v}$ (or $\widetilde{\textbf{F}_n}$) can be regarded as the modality-relevant information, which can be written as follows:
\begin{small}
\begin{equation}
\begin{array}{c} 
  \textbf{M}_v = \textbf{F}_v - \widetilde{\textbf{F}}_v, \quad \textbf{M}_n = \textbf{F}_n - \widetilde{\textbf{F}}_n.
\end{array}
\end{equation}
\end{small}

Although $\textbf{M}_v$ and $\textbf{M}_n$ can reflect the modality information, the IN layers may cause that some discriminative information is discarded. Moreover, there is still some modality-irrelevant information in $\textbf{M}_v$ and $\textbf{M}_n$. Therefore, we further distill it through the proposed MRM and MCM to obtain both modality-irrelevant and modality-relevant information.

\subsubsection{Modality Restitution Module.} 

The Modality Restitution Module (MRM) is designed for restituting modality-irrelevant but identity-relevant information to the network to ensure high performance of our MRCN.

Since the modality discrepancy between the VIS and NIR modalities mainly lies in the channel space \cite{li2020infrared, zhang2021towards}, the proposed MRM adopts two simple channel attention modules $\mathcal{A}_{v1}$ and $\mathcal{A}_{n1}$ to distill the modality-irrelevant but identity-relevant information from $\textbf{M}_v$ and $\textbf{M}_n$, respectively. Then, we obtain the distilled information $\textbf{M}_{v+}$ and $\textbf{M}_{n+}$, which can be written as follows:
\begin{small}
\begin{equation}
\begin{array}{c} 
  \textbf{M}_{v+} = \textbf{M}_v \times \mathcal{A}_{v1}(\textbf{M}_v), \quad \textbf{M}_{n+} = \textbf{M}_n \times \mathcal{A}_{n1}(\textbf{M}_n),
\end{array}
\end{equation}
\end{small}where the channel attention modules $\mathcal{A}_{v1}$ and $\mathcal{A}_{n1}$ adopt SE-Net \cite{hu2018squeeze}, which is composed of a global average pooling layer and two fully-connected layers, followed by a ReLU activation function and a sigmoid activation function. To reduce the number of parameters, the dimension reduction ratio is set to 16.

Then, we restitute the modality-irrelevant but identity-relevant information $\textbf{M}_{v+}$ and $\textbf{M}_{n+}$ obtained through the above distillation process to the normalized modality feature $\textbf{F}_{v+}$ and $\textbf{F}_{n+}$, which can be written as follows:
\begin{small}
\begin{equation}
\begin{array}{c} 
  \textbf{F}_{v+} = \widetilde{\textbf{F}}_v + \textbf{M}_{v+}, \quad \textbf{F}_{n+} = \widetilde{\textbf{F}}_n + \textbf{M}_{n+}.
\end{array}
\end{equation}
\end{small}

Finally, $\textbf{F}_{v+}$ and $\textbf{F}_{n+}$ are used as the output features of MRM to optimize the network to ensure the high performance of the proposed MRCN.

\subsubsection{Modality Compensation Module.} 

The Modality Compensation Module (MCM) is designed to compensate modality-relevant but identity-irrelevant information for the other modality to reduce the modality discrepancy between the two modalities.

Similar to MRM, we also employ two SENet-like channel attention modules $\mathcal{A}_{v2}$ and $\mathcal{A}_{n2}$ to distill the difference $\textbf{M}_v (\textbf{M}_n)$ between the original input feature $\textbf{F}_v$ (or $\textbf{F}_n$) and the normalized modality feature $\widetilde{\textbf{F}_n}$ (or $\widetilde{\textbf{F}_v} $). Then, we can obtain the modality-relevant but identity-irrelevant information $\textbf{M}_{v-}$ and $\textbf{M}_{n-}$, which can be formulated as follows:
\begin{small}
\begin{equation}
\begin{array}{c} 
  \textbf{M}_{v-} = \textbf{M}_v \times \mathcal{A}_{v2}(\textbf{M}_v), \quad \textbf{M}_{n-} = \textbf{M}_n \times \mathcal{A}_{n2}(\textbf{M}_n).
\end{array}
\end{equation}
\end{small}

\begin{figure}[t]
\includegraphics[height=3.5cm,width=8.3cm]{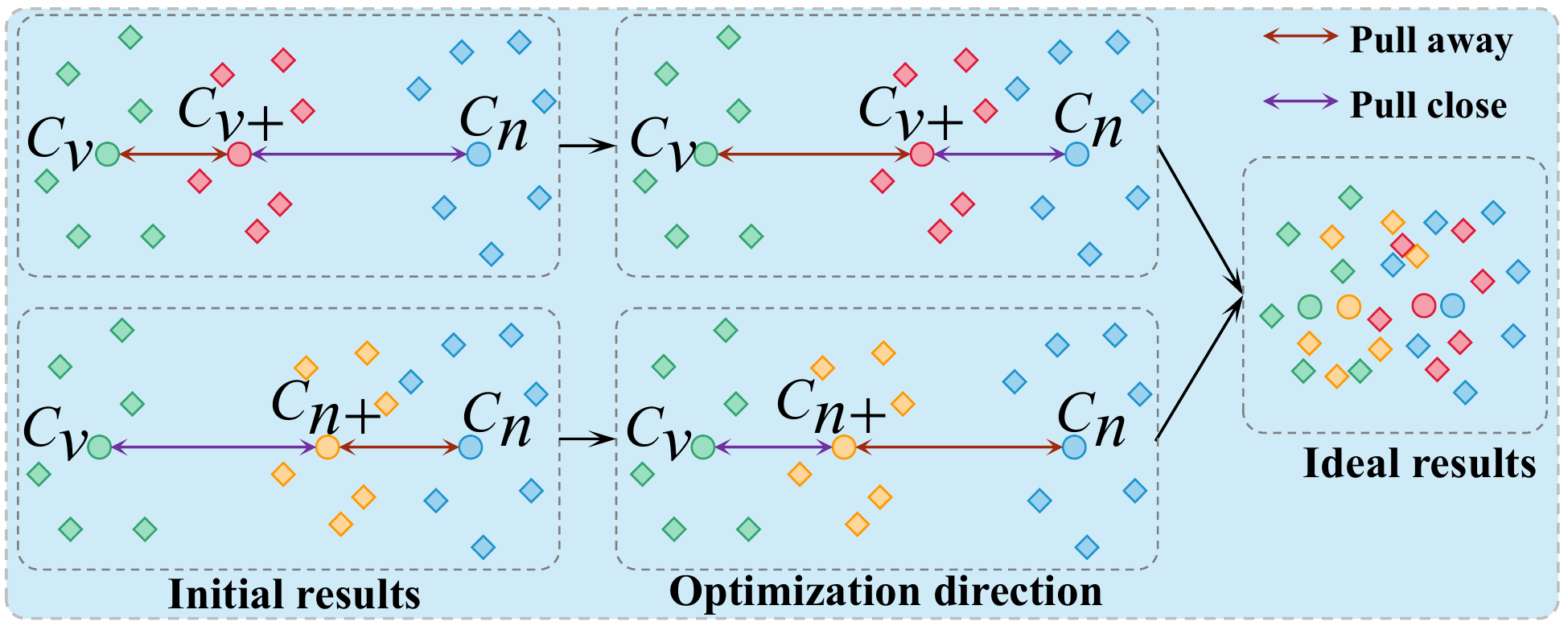}
\caption{Illustration of the proposed CQC loss (taking the feature centers $\textbf{c}_{v+}$ and $\textbf{c}_{n+}$ output by MRM as an example). Different colors correspond to different features from the original VIS features $\textbf{f}_v$ and NIR features $\textbf{f}_n$, and the features $\textbf{f}_{v+}$ and $\textbf{f}_{n+}$ output by MRM. Circles denote the class centers of different features.}
\label{img:img3}
\end{figure}

The difference between MCM and MRM is that we compensate for the modality-relevant but identity-irrelevant information for its counterpart modality. Here, let $\textbf{F}_{v-}$ and $\textbf{F}_{n-}$ respectively denote the compensated features distilled by MCM in the VIS and NIR modalities. Then we have:
\begin{small}
\begin{equation}
\begin{array}{c} 
  \textbf{F}_{v-} = \widetilde{\textbf{F}}_v + \textbf{M}_{n-}, \quad \textbf{F}_{n-} = \widetilde{\textbf{F}}_n + \textbf{M}_{v-}.
\end{array}
\end{equation}
\end{small}

In this way, we can obtain modality-relevant but identity-irrelevant information from the above distilltion process. By adding $\textbf{M}_{v-}$ (or $\textbf{M}_{n-}$) to the normalized modality feature $\textbf{F}_n$ (or $\textbf{F}_v$), we can obtain compensated features $\textbf{F}_{n-}$ (or $\textbf{F}_{v-}$), which can be regarded as the corresponding features of a person in the VIS (or NIR) modality. Therefore, the output features $\textbf{F}_{v-}$ and $\textbf{F}_{n-}$ of MCM are beneficial to jointly optimize the network. By this way, the modality differences between the VIS modality and the NIR modality can be significantly reduced.

\subsection{Center-Quadruplet Causal Loss}
To better disentangle the modality-relevant and modality-irrelevant information, we propose a new Center-Quadruplet Causal (CQC) loss to constrain the output features to enable our network to reduce the modality discrepancy while maintaining high performance. We denote the features output by MRCN as $\textbf{\textit{{f}}} = \left\{{\textbf{f}_v, \textbf{f}_{v+}, \textbf{f}_{v-}, \textbf{f}_{n-}, \textbf{f}_{n+}, \textbf{f}_n}\right\}$, where $\textbf{f}_v$ and $\textbf{f}_n$ are the original features of the VIS and NIR modalities, $\textbf{f}_{v+}$ and $\textbf{f}_{n+}$ are the features output by MRM, and $\textbf{f}_{v-}$ and $\textbf{f}_{n-}$ are the features output by MCM.

For MRM, the proposed CQC loss aims to make the features output by MRM more discriminative. Specifically, we first compute the feature centers of the restituted features ($\textbf{f}_{v+} / \textbf{f}_{n+}$) for each identity. We denote the center features as ($\textbf{c}_{v+} / \textbf{c}_{n+}$). Then, we make the feature centers ($\textbf{c}_{v+} / \textbf{c}_{n+}$) have less distance from the original feature center ($\textbf{c}_n / \textbf{c}_v$) of the other modality than its distance from the original feature center ($\textbf{c}_v / \textbf{c}_n$) of the same identity. Thus, for MRM, the proposed CQC loss can be formulated as follows:
\begin{small}
\begin{equation}
\begin{split}
{\mathcal{L}^{(\textbf{c}_v, \textbf{c}_n, \textbf{c}_{v+}, \textbf{c}_{n+})}_{MRM}} &= \sum\limits_{i=1}^C [ \alpha  +
\mathop{\textbf{\textit{D}}}(\textbf{c}^{i}_{v+}, \textbf{c}^{i}_n) - \mathop{\textbf{\textit{D}}}(\textbf{c}^{i}_{v+}, \textbf{c}^{i}_v)]_+ \\
&+ \sum\limits_{i=1}^C [ \alpha  +
\mathop{\textbf{\textit{D}}}(\textbf{c}^{i}_{n+}, \textbf{c}^{i}_v) - \mathop{\textbf{\textit{D}}}(\textbf{c}^{i}_{n+}, \textbf{c}^{i}_n)]_+,
\end{split}
\end{equation}
\end{small}
where ${\textbf{\textit{D}}}(\textbf{a}, \textbf{b})$ is the Euclidean distance between the feature $\textbf{a}$ and the feature $\textbf{b}$, and $C$ is the class size in a minibatch, $\alpha$ is a margin parameter and $[z]_{+} = max(z,0)$. $\textbf{c}^{i}_v$, $\textbf{c}^{i}_n$, $\textbf{c}^{i}_{v+}$ and $\textbf{c}^{i}_{n+}$ are from the same identity.

\begin{table*}
  \centering
  \tabcolsep=0.065cm
  \renewcommand{\arraystretch}{1.3}
  \fontsize{10pt}{10pt}\selectfont
  \begin{tabular}{l|cccccccc|ccccccccc}
  \hline
   \multirow{3}{*}{Model} & \multicolumn{8}{c|}{RegDB} & \multicolumn{8}{c}{SYSU-MM01} \\
   \cline{2-17}
   &\multicolumn{4}{c}{Visible to Infrared} &\multicolumn{4}{c|}{Infrared to Visible} &\multicolumn{4}{c}{All Search} &\multicolumn{4}{c}{Indoor Search}\\
   \cline{2-17}
   & R-1 & R-10 & R-20 & mAP   & R-1 & R-10 & R-20 & mAP & R-1 & R-10 & R-20 & mAP & R-1 & R-10 & R-20 & mAP\\
   \hline
    D$^{2}$RL\cite{wang2019learning}   & 43.4 & 66.1 & 76.3 & 44.1 & -    & -    & -    & -    & 28.9 & 70.6 & 82.4 & 29.2 & -    & -    & -    & - \\
    JSIA-ReID\cite{wang2020cross}      & 48.1 & -    & -    & 48.9 & 48.5 & -    & -    & 49.3 & 38.1 & 80.7 & 89.9 & 36.9 & 43.8 & 86.2 & 94.2 & 52.9 \\
    AlignGAN\cite{wang2019aligngan}     & 57.9 & -    & -    & 53.6 & 56.3 & -    & -    & 53.4 & 42.4 & 85.0 & 93.7 & 40.7 & 45.9 & 87.6 & 94.4 & 54.3 \\
    X-Modality\cite{li2020infrared}    & 62.2 & 83.1 & 91.7 & 60.2 & -    & -    & -    & -    & 49.9 & 89.8 & 96.0 & 50.7 & -    & -    & -    & - \\
    Hi-CMD\cite{choi2020hi}            & 70.9 & 86.4 & -    & 66.0 & -    & -    & -    & -    & 34.9 & 77.6 & -    & 35.9 & -    & -    & -    & - \\
    MCLNet\cite{hao2021cross}            & 80.3 & 92.7 & 96.0 & 73.1 & 75.9 & 90.9 & 94.6 & 69.5 & 65.4 & 93.3 & 97.1 & 62.0 & 72.6 & 97.0 & 99.2 & 76.6\\
    CM-NAS\cite{fu2021cm}              & 82.8 &95.1 &97.7 & 79.3 & 81.7 &94.1 &96.9 & 77.6 & 60.8 & 92.1 & 96.8 & 58.9 & 68.0 & 94.8 & 97.9 & 52.4 \\
    DART\cite{yang2022learning} & 83.6 & -    & - & 75.7 & 82.0 & -    & - & 73.8 & 68.7 & 96.4 & 99.0 & 66.3 & 72.5 & 97.8 & 99.5 & 78.2\\
 
    SMCL\cite{wei2021syncretic}         &83.9 & -    & -    &79.8 &83.1 & -    & -    &78.6 &67.4 & 92.9 & 96.8 & 61.8 & 68.8 & 96.6 & 98.8 & 75.6  \\
    CAJ\cite{ye2021channel} & 85.0 & 95.5 & 97.5 & 79.1 & 84.8 & 95.3 & 97.5 & 77.8  & 69.9 & 95.7 & 98.5 & 66.9 & 76.3 & 97.9 & 99.5 & 80.4\\
    FMCNet\cite{zhang2022fmcnet} & 89.1 & -    & - & 84.4 & \underline{88.4} & -    & - & \underline{83.9} & 66.3 & -    & - & 62.5 & 68.2 & -    & - & 74.1 \\
    
    \textbf{Our MRCN}               & 91.4 & \underline{98.0} & \underline{99.0} & \underline{84.6} & 88.3 & \underline{96.7} & \underline{98.5} & 81.9 & 68.9 & 95.2 & 98.4 & 65.5 & 76.0 & \underline{98.3} & \underline{99.7} & 79.8     \\
    \hline

    NFS\cite{chen2021neural}             & 80.5 & 91.6 & 95.1 & 72.1 &78.0  & 90.5 & 93.6 & 69.8 & 56.9 & 91.3 & 96.5 & 55.5 & 62.8 & 96.5 & 99.1 & 69.8\\
    DDAG\cite{ye2020dynamic}            & 69.3 & 86.2 & 91.5 & 63.5 & 68.1 & 85.2 & 90.3 & 61.8 & 54.8 & 90.4 & 95.8 & 53.0 & 61.0 & 94.1 & 98.4 & 68.0\\
    MPANet\cite{wu2021discover} & 83.7 & - & - & 80.9 & 82.8 & - & - & 80.7 & \underline{70.6} & \underline{96.2} & 98.8 & \underline{68.2} & \underline{76.7} & 98.2 & 99.6 & \textbf{81.0} \\
    MMN\cite{zhang2021towards} &  \underline{91.6} & 97.7 & 98.9 & 84.1 & 87.5 & 96.0 & 98.1 & 80.5 & \underline{70.6} & \underline{96.2} & \underline{99.0} & 66.9 & 76.2 & 97.2 & 99.3 & 79.6 \\
     
    \textbf{Our MRCN-P}               & \textbf{95.1} & \textbf{98.8} & \textbf{99.5} & \textbf{89.2} & \textbf{92.6} & \textbf{98.0} & \textbf{99.1} & \textbf{86.5} & \textbf{70.8} & \textbf{96.5} & \textbf{99.1} & \textbf{67.3} & \textbf{76.4} & \textbf{98.5} & \textbf{99.9} & \underline{80.0}    \\
    \hline
    \end{tabular}
    \caption{Performance obtained by the competing methods on the RegDB and SYSU-MM01 datasets. We divide the competing methods into two groups:  the global-feature-based methods (top row) and the part-feature-based methods (bottom row).}
    \label{tab:tab1}
\end{table*}

Similarly, for MCM, the proposed CQC loss should enable the features output by MCM to reduce the modality gaps between the VIS and NIR images. Specifically, we first compute the feature centers of the compensated features ($\textbf{f}_{v-} / \textbf{f}_{n-}$) for each identity. We set the center features as ($\textbf{c}_{v-} / \textbf{c}_{n-}$). Then, we make the center features ($\textbf{c}_{v-} / \textbf{c}_{n-}$) have less distance from the feature centers ($\textbf{c}_n / \textbf{c}_v$) of the other modality than its distance from the feature centers ($\textbf{c}_v / \textbf{c}_n$) of the original modality for the same identity. Thus, for MCM, the proposed CQC loss can be written as follows:
\begin{small}
\begin{equation}
\begin{split}
{\mathcal{L}^{(\textbf{c}_v, \textbf{c}_n, \textbf{c}_{v-}, \textbf{c}_{n-})}_{MCM}} &= \sum\limits_{i=1}^C {[\alpha + \mathop{\textbf{\textit{D}}}(\textbf{c}^{i}_{v-}, \textbf{c}^{i}_n)} - \mathop{\textbf{\textit{D}}}(\textbf{c}^{i}_{v-}, \textbf{c}^{i}_v)]_{+} \\
&+ \sum\limits_{i=1}^C {[ \alpha  +
\mathop{\textbf{\textit{D}}}(\textbf{c}^{i}_{n-}, \textbf{c}^{i}_v)} - \mathop{\textbf{\textit{D}}}(\textbf{c}^{i}_{n-}, \textbf{c}^{i}_n)]_{+}.
\end{split}
\end{equation}
\end{small}

Then, all the CQC loss function is formulated as follows:
\begin{small}
\begin{equation}
{\mathcal{L}_{CQC}} = {\mathcal{L}^{(\textbf{c}_v, \textbf{c}_n, \textbf{c}_{v+}, \textbf{c}_{n+})}_{MRM}} + {\mathcal{L}^{(\textbf{c}_v, \textbf{c}_n, \textbf{c}_{v-}, \textbf{c}_{n-})}_{MCM}}.
\end{equation}
\end{small}

\subsection{Multi-Loss Optimization}

Besides the proposed ${\mathcal{L}_{CQC}}$, we also combine the label-smoothed cross-entropy loss ${\mathcal{L}_{lsce}}$ \cite{Luo2019Bags} and the triplet loss ${\mathcal{L}_{tri}}$ \cite{hermans2017defense} to jointly optimize the network by minimizing the sum ${\mathcal{L}_{total}}$ of these three losses, which can be formulated as follows:
\begin{small}
\begin{equation}
\label{eq1}
{\mathcal{L}_{total}} = {\mathcal{L}_{lsce}} + \lambda_1{\mathcal{L}_{tri}} + \lambda_2{\mathcal{L}_{CQC}},
\end{equation}
\end{small}where $\lambda_1$ and $\lambda_2$ are the coefficients to control the relative importance of the loss terms.

\section{Experiments}
\subsection{Datasets}
The SYSU-MM01 dataset \cite{wu2017rgb} contains 491 identities captured by 4 VIS cameras and 2 NIR cameras. The training set contains $19,659$ VIS images and $12,792$ NIR images of $395$ identities, and the testing set contains $3,803$ NIR images of $96$ identities as the query set. The RegDB dataset \cite{nguyen2017person} consists of 412 identities, and each identity has 10 VIS images and 10 NIR images captured by a pair of overlapping cameras. Following \cite{ye2018hierarchical}, we evaluate the competing methods in both Visible to Infrared and Infrared to Visible modes.

\subsection{Implementation Details}
All the input images are resized to $3 \times 288 \times 128$, and the random horizontal flip and random erasing \cite{zhong2020random} techniques are adopted for data augmentation during the training phase. 
The initial learning rate is set to $1\times10^{-2}$ and then it linearly increases to $1\times10^{-1}$ after 10 epochs. After the warm-up process, we decay the learning rate to $1\times10^{-2}$ at 20 epoch, and further decay to $1\times10^{-3}$ at epoch 60 until a total of 80 epochs. In each mini-batch, we randomly select 4 VIS images and 4 NIR images of 4 identities for training. The SGD optimizer is adopted for optimization, and the momentum parameter is set to 0.9. For the margin parameter in the CQC loss, we experimentally set it to 0.2. For the coefficients $\lambda_1$ in Eq. (\ref{eq1}), we set it to 1.

\subsection{Comparison with State-of-the-Art Methods}

We first compare our method with several state-of-the-art methods to demonstrate the superiority of our method.

\textbf{RegDB:} In Tab. \ref{tab:tab1}, we can see that the experimental results on RegDB show that MRCN achieves the best performance under different testing modes against all the competing state-of-the-art methods. Specifically, for the Visible to Infrared mode, MRCN (MRCN-P) achieves 91.4\% (95.1\%) in Rank-1 accuracy and 84.6\% (89.2\%) in mAP. MRCN-P outperforms the second best MMN \cite{zhang2021towards} by 3.5\% in Rank-1 accuracy and 5.1\% in mAP, respectively. For the Infrared to Visible mode, MRCN (MRCN-P) also obtains 88.3\% (92.6\%) in Rank-1 accuracy and 81.9\% (86.5\%) in mAP. MRCN-P outperforms the second best method MMN \cite{zhang2021towards} by 3.5\% in Rank-1 accuracy and 5.1\% in mAP, respectively. 

\textbf{SYSU-MM01:} The results on SYSU-MM01 in Tab. \ref{tab:tab1} show that MRCN obtains competitive performance under both All-Search and Indoor-Search modes. For the All-Search mode, MRCN (MRCN-P) achieves 68.9\% (70.8\%) in Rank-1 accuracy and 65.5\% (67.3\%) in mAP. For the Indoor-Search mode, MRCN (MRCN-P) achieves 76.0\% (76.4\%) in Rank-1 accuracy and 79.8\% (80.0\%) in mAP. The comparative results validate the effectiveness of our method. Moreover, the results also demonstrate that MRCN (MRCN-P) can effectively reduce the modality discrepancy between the VIS and NIR images.

Moreover, compared with SYSU-MM01, person pose in RegDB is more aligned between the VIS and NIR images. The MCM in MRCN is used to compensate modality-relevant features for the features of the other modality, which has a positive impact when the pose is aligned. Thus, MRCN can yield better results on RegDB. As for the reasoning efficiency, on the one hand, although MRCN consumes more time than these methods, the evaluation of VI-ReID task does not consume much time (about 25-32s). On the other hand, although using more time, MRCN achieves significant result improvements (95.1\% in Rank-1 and 89.2\% in mAP).

\subsection{Ablation Studies}

\textbf{The influence of different components.} To demonstrate the contribution of each component to MRCN, we conduct some ablation studies on RegDB and SYSU-MM01. As shown in Tab. \ref{tab:tab2}, MRCN without using the CQC loss can improve the performance of the baseline model, which indicates the modality restitution and compensation can effectively reduce modality differences. In constrast, MRCN with the CQC loss can facilitate the decoupling of modality-relevant and modality-irrelevant features. Besides, the pull-away of features helps to effectively learn more abundant information, which in turn improves the model performance. Moreover, both MRM and MCM can improve the performance of the baseline model, and the combination of the two components can achieve the best performance, indicating that MRM and MCM can complement to each other.

\textbf{Effectiveness on which stage of ResNet-50 to plug MRCN into.} We plug MRCN into different stages of ResNet-50 to study how it will affect the performance of MRCN. As shown in Tab. \ref{tab:tab3}, MRCN after stage-1 on RegDB and MRCN after stage-0 on SYSU-MM01 can achieve the best performance, respectively, which indicates that after stage-1 for RegDB and after stage-0 for SYSU-MM01, the proposed MRCN is more suitable for the separation of modality-relevant and modality-irrelevant features.

\begin{table}[t]
  \centering
  \renewcommand{\arraystretch}{1.15}
  \begin{tabular}{l|cc|cc}\hline
    \multirow{2}{*}{Methods} & \multicolumn{2}{c|}{RegDB}& \multicolumn{2}{c}{SYSU-MM01} \\
    \cline{2-5}
                         & R-1 & mAP & R-1 & mAP  \\\hline
baseline          & 78.2	& 70.4  & 60.7	& 57.7 \\ 
\hline
MRCN (w/o CQC)     & 85.5	& 79.0   & 64.6	& 61.6 \\
MCM (w CQC)               & 89.7	& 83.1   & 64.4	& 61.2 \\
MRM (w CQC)               & 90.3	& 83.4   & 66.8	& 63.3  \\
\textbf{MRCN}& \textbf{91.4} & \textbf{84.6} & \textbf{68.9} & \textbf{65.5}\\ 
    \hline
    \end{tabular}
  \caption{Influence of different components on MRCN.}
    \label{tab:tab2}
\end{table}

\begin{table}[t]
  \centering
  \renewcommand{\arraystretch}{1.15}
  \begin{tabular}{l|cc|cc}\hline
    \multirow{2}{*}{Methods} & \multicolumn{2}{c|}{RegDB}& \multicolumn{2}{c}{SYSU-MM01} \\    \cline{2-5}
                & R-1 & mAP & R-1 & mAP  \\
   \hline
MRCN after stage-0	& 87.8	& 80.4  & \textbf{68.9} & \textbf{65.5}\\ 
\textbf{MRCN after stage-1}& \textbf{91.4} & \textbf{84.6} &68.3	& 65.4\\
MRCN after stage-2	& 88.1	 & 80.1  & 63.4	& 59.4 \\ 
MRCN after stage-3	& 73.5 & 67.5  & 58.2	& 56.4 \\ 
MRCN after stage-4	& 72.0	& 67.8  & 59.6	& 57.7\\ 
    \hline
    \end{tabular}
  \caption{Effectivenes on which stage of ResNet-50 to plug MRCN into.}
    \label{tab:tab3}
\end{table}

\begin{table}[t]
  \centering
  \renewcommand{\arraystretch}{1.15}
  \begin{tabular}{l|cc|cc}\hline
    \multirow{2}{*}{Methods} & \multicolumn{2}{c|}{RegDB}& \multicolumn{2}{c}{SYSU-MM01} \\    \cline{2-5}
                         & R-1 & mAP & R-1 & mAP  \\
   \hline
baseline	& 78.2	& 70.4	& 60.7	& 57.7  \\ \hline
SNR	& 85.1	& 77.4	& 63.8	& 61.2  \\ 
\textbf{MRCN}	&\textbf{91.4} & \textbf{84.6} & \textbf{68.9} & \textbf{65.5}\\ 
    \hline
    \end{tabular}
  \caption{Comparison with SNR under the same baseline.}
    \label{tab:tab4}
\end{table}

\begin{figure}[t]
\centering
\includegraphics[height=6.6cm,width=8.5cm]{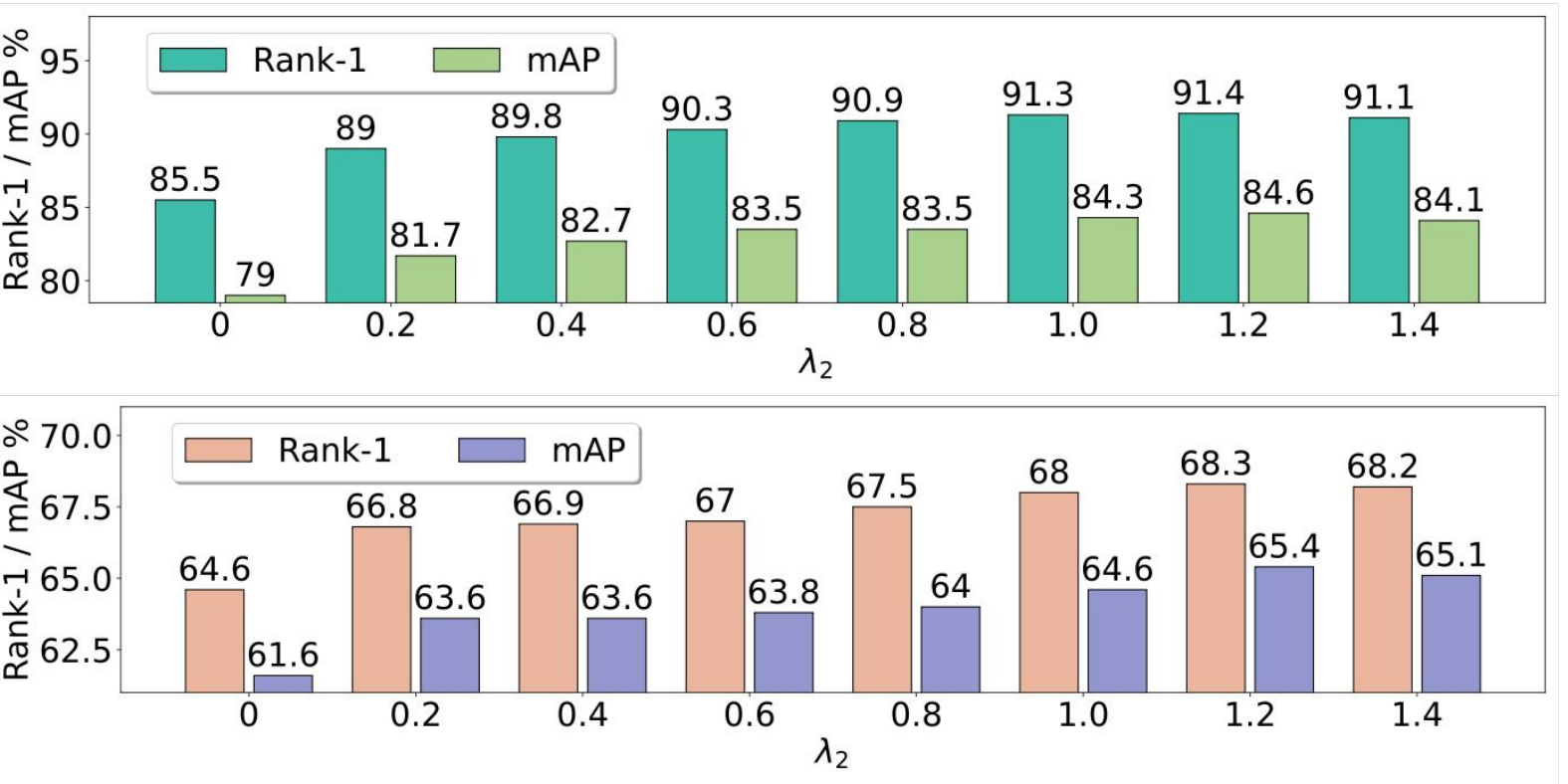}
\caption{Influence of different $\lambda_2$ in Eq. (\ref{eq1}) on the RegDB (top row) and SYSU-MM01 datasets (bottom row).}
\label{img:img4}
\end{figure}

\begin{figure*}[t]
\centering
\includegraphics[height=5.8cm,width=17.6cm]{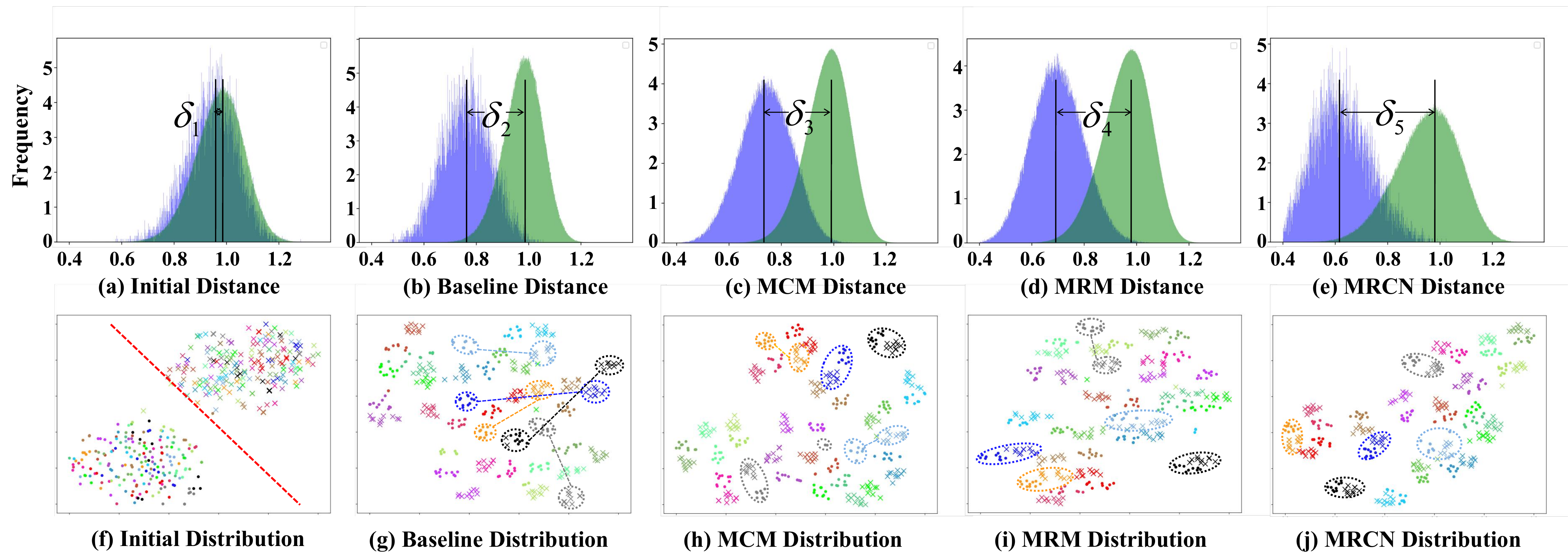}
\caption{(a-e) The frequency of intra-class and inter-class distances between the cross-modality features of SYSU-MM01, which are indicated by the blue and green colors, respectively. (f-j) The features distribution in the 2D feature space. A total of 20 persons are selected from the test set. The samples with the same color are from the same person. The “dot” and “cross” markers denote the images from the VIS and the NIR modalities, respectively.}
\label{img:img6}
\end{figure*}

\textbf{Comparison with SNR.} The SNR \cite{jin2020style} is similar to MRCN because that both methods adopt a feature disentanglement technique to reduce the style gaps. However, MRCN differs from SNR in the following three aspects: (1) SNR only feeds the identity-relevant features to the network while discarding the identity-irrelevant ones. However, MRCN compensates that for the features of the other modality. (2) The output of SNR is directly input into the loss for decoupling while MRCN inputs all the features into the rest of the backbone, which can be viewed as further distillation and thus it is more conducive for CQC loss to decouple modality information. That is the reason why MRCN outperforms SNR. (3) SNR is suitable for small style gaps which can be discarded. However, when faced with large style gaps such as VIS and NIR images, it is better to compensate the style gaps for the other modality. Hence, MRCN outperforms SNR in VI-ReID.
For a fair comparison, we conduct experiments using the same baseline for both SNR and MRCN in Tab. \ref{tab:tab4}, which shows that MRCN outperforms SNR by 6.3\% in Rank-1 accuracy and 7.2\% in mAP on RegDB, and by 5.1\% in Rank-1 accuracy and 4.3\% in mAP on SYSU-MM01, respectively. Experimental results indicate that MRCN is more effective than SNR in reducing the modality differences.

\textbf{The influence of different attention blocks.} In MRCN, the SE-block is used to distill the modality information, and other attention blocks can also achieve this goal. The key to the performance improvement of MRCN is to input the features into the rest of the backbone for decoupling them to through the CQC loss. The results of MRCN with/without the CQC loss in Tab. \ref{tab:tab2} confirm this. Moreover, we compare the influence of different attention blocks (CBAM \cite{woo2018cbam}, ECA-block \cite{wang2020eca}, SE-block) on MRCN on RegDB (R-1 / mAP) as follows: 91.1 / 83.9, 91.4 / 84.1, 91.4 / 84.6. It shows that different attention blocks has no significant influence on the performance.

\textbf{The influence of hyperparameter $\lambda_2$ in the CQC loss.} In Eq. (\ref{eq1}), we use a parameter $\lambda_2$ to control the trade-off between $\mathcal{L}_{CQC}$ with the $\mathcal{L}_{lsce}$ and $\mathcal{L}_{tri}$. To evaluate the influence, we give quantitative comparisons and report the results in Fig. \ref{img:img4}. From the results, the best performance is achieved when the parameter $\lambda_2$ is set to 1.2.

\begin{figure}[t]
\includegraphics[height=5.4cm,width=7.8cm]{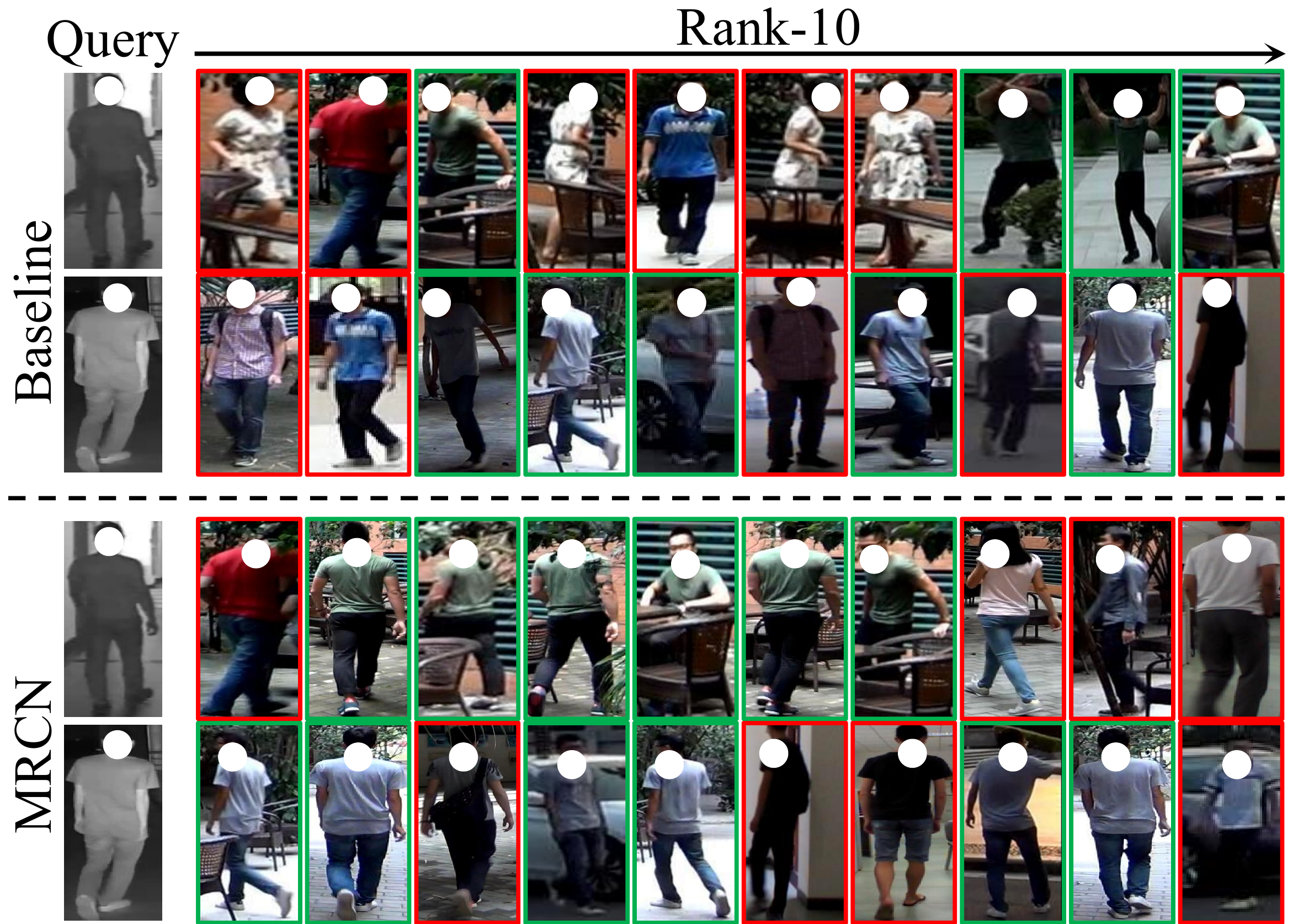}
\caption{The Rank-10 retrieval results obtained by the baseline and our MRCN on the SYSU-MM01 dataset.}
\label{img:img7}
\end{figure}

\subsection{Visualization}

\textbf{Feature distribution.} To investigate the reason why MRCN is effective, we conduct experiments to compute the frequency of inter-class and intra-class distances in Fig. \ref{img:img6} (a-e). By comparing Fig. \ref{img:img6} (c-e) with Fig. \ref{img:img6} (a-b), the means of inter-class and intra-class distances are pushed away by MRM and MCM, where $\delta_1$ \textless $\delta_2$ \textless $\delta_3$ and  $\delta_1$ \textless $\delta_2$ \textless $\delta_4$. Besides, the intra-class distance of MRCN is significantly reduced compared with the intra-class distance of the initial features and the baseline features. It shows that MRCN can effectively reduce the modality discrepancy between the VIS modality and the NIR modality.
For further validating the effectiveness of the proposed MRCN, we plot the t-SNE \cite{laurens2008Visualizing} distribution of the MRCN feature representations in the 2D feature space for visualization. As shown in Fig. \ref{img:img6} (f-j), the proposed MRCN can greatly shorten the distances between the images corresponding to the same identity in the VIS modality and the NIR modality, and effectively reduce the modality discrepancy.

\textbf{Retrieval result.} 
To further evaluate the proposed MRCN, we compare the retrieval results obtained by our method with those obtained by the baseline on several image pairs of the SYSU-MM01 dataset, using the multi-shot setting and the all-search mode. The results are shown in Fig. \ref{img:img7}. For each retrieval case, the query images shown in the first column are the NIR images, and the gallery images shown in the following columns are the VIS images. The retrieved images with the green bounding boxes belong to the same identities as the query, while the images with the red bounding boxes are opposite to the query. In general, the proposed MRCN can effectively improve the ranking results with more green bounding boxes ranked in the top positions.

\section{Conclusion}

In this paper, we propose a novel MRCN to narrow the gap between the VIS and NIR modalities. Specifically, we first reduce the modality discrepancy by using two IN layers. Next, to reduce the influence of IN layers on removing discriminative information, we propose a MRM module to distill modality-irrelevant features and propose a MCM module to distill modality-relevant features from the removed information. Then, the modality-irrelevant features are used to restitute to the normalized VIS and NIR features, while the modality-relevant features are used to compensate for the features of the other modality.
Furthermore, we propose a new CQC loss to encourage the network to effectively learn the disentangled features. Extensive experiments are conducted to validate the superiority of our method for VI-ReID on the challenging SYSU-MM01 and RegDB datasets. 

\section{Acknowledgments}

This work was supported by the National Key Research and Development Program of China under Grant 2022ZD0160402, by the FuXiaQuan National Independent Innovation Demonstration Zone Collaborative Innovation Platform Project under Grant 3502ZCQXT2022008, by the National Natural Science Foundation of China under Grants U21A20514, 62176195 and 62071404, and by the Open Research Projects of Zhejiang Lab under Grant 2021KG0AB02.

\bibliography{aaai23}

\end{document}